\documentclass[letterpaper]{article} 
\usepackage{aaai23}  
\usepackage{times}  
\usepackage{helvet}  
\usepackage{courier}  
\usepackage[hyphens]{url}  
\usepackage{graphicx} 
\usepackage{natbib}  
\usepackage{caption} 
\usepackage{algorithm}
\usepackage{algorithmic}

\usepackage{newfloat}
\usepackage{listings}
\DeclareCaptionStyle{ruled}{labelfont=normalfont,labelsep=colon,strut=off} 
\floatstyle{ruled}
\newfloat{listing}{tb}{lst}{}
\floatname{listing}{Listing}
\title{Entity-Agnostic Representation Learning for Parameter-Efficient \protect\\Knowledge Graph Embedding}

\author{
    Mingyang Chen\textsuperscript{\rm 1}\equalcontrib,
    Wen Zhang\textsuperscript{\rm 2}\equalcontrib, 
    Zhen Yao\textsuperscript{\rm 2},
    Yushan Zhu\textsuperscript{\rm 1},
    Yang Gao\textsuperscript{\rm 4}, \\
    Jeff Z. Pan\textsuperscript{\rm 5},
    Huajun Chen\textsuperscript{\rm 1,3,6}\thanks{Corresponding Author.}
}
\affiliations{
    \textsuperscript{\rm 1}College of Computer Science and Technology, Zhejiang University\\
    \textsuperscript{\rm 2}School of Software Technology, Zhejiang University \\
    \textsuperscript{\rm 3}Donghai Laboratory 
    \textsuperscript{\rm 4}Huawei Technologies Co., Ltd. \\
    \textsuperscript{\rm 5}School of Informatics, The University of Edinburgh \\
    \textsuperscript{\rm 6}Alibaba-Zhejiang University Joint Institute of Frontier Technologies \\
    \{mingyangchen, zhang.wen, yz0204, yushanzhu, huajunsir\}@zju.edu.cn, \\
    frank.gaoyang@huawei.com, j.z.pan@ed.ac.uk
}

\usepackage{bibentry}

\usepackage{amssymb}
\usepackage{amsmath}
\usepackage{booktabs}
\usepackage{enumitem}
\usepackage{multirow}
\newcommand{\model}{EARL}
\usepackage{colortbl}

\usepackage{xcolor}

\begin{document}

\maketitle

\begin{abstract}
We propose an entity-agnostic representation learning method for handling the problem of inefficient parameter storage costs brought by embedding knowledge graphs. 
Conventional knowledge graph embedding methods map elements in a knowledge graph, including entities and relations, into continuous vector spaces by assigning them one or multiple specific embeddings  (i.e., vector representations). Thus the number of embedding parameters increases linearly as the growth of knowledge graphs.  
In our proposed model, \underline{E}ntity-\underline{A}gnostic \underline{R}epresentation \underline{L}earning (\model), we only learn the embeddings for a small set of entities and refer to them as reserved entities. To obtain the embeddings for the full set of entities, we encode their distinguishable information from their connected relations, $k$-nearest reserved entities, and multi-hop neighbors. We learn universal and entity-agnostic encoders for transforming distinguishable information into entity embeddings.
This approach allows our proposed \model~to have a static, efficient, and lower parameter count than conventional knowledge graph embedding methods.
Experimental results show that \model~uses fewer parameters and performs better on link prediction tasks than baselines, reflecting its parameter efficiency.
\end{abstract}

\section{Introduction}

Recently, many knowledge graphs (KGs)~\cite{PVGW2017}, including Freebase \cite{freebase}, NELL \cite{nell}, Wikidata \cite{wikidata}, and YAGO \cite{yago4} have been used as the knowledge resource for a myriad of applications in the field of natural language processing \cite{LM-KG-1, LM-KG-2}, as well as in the study of computer vision \cite{ImgCap-KG-1, VQA-KG-1}.

\begin{figure}[t]
\centering
\includegraphics[scale=0.28]{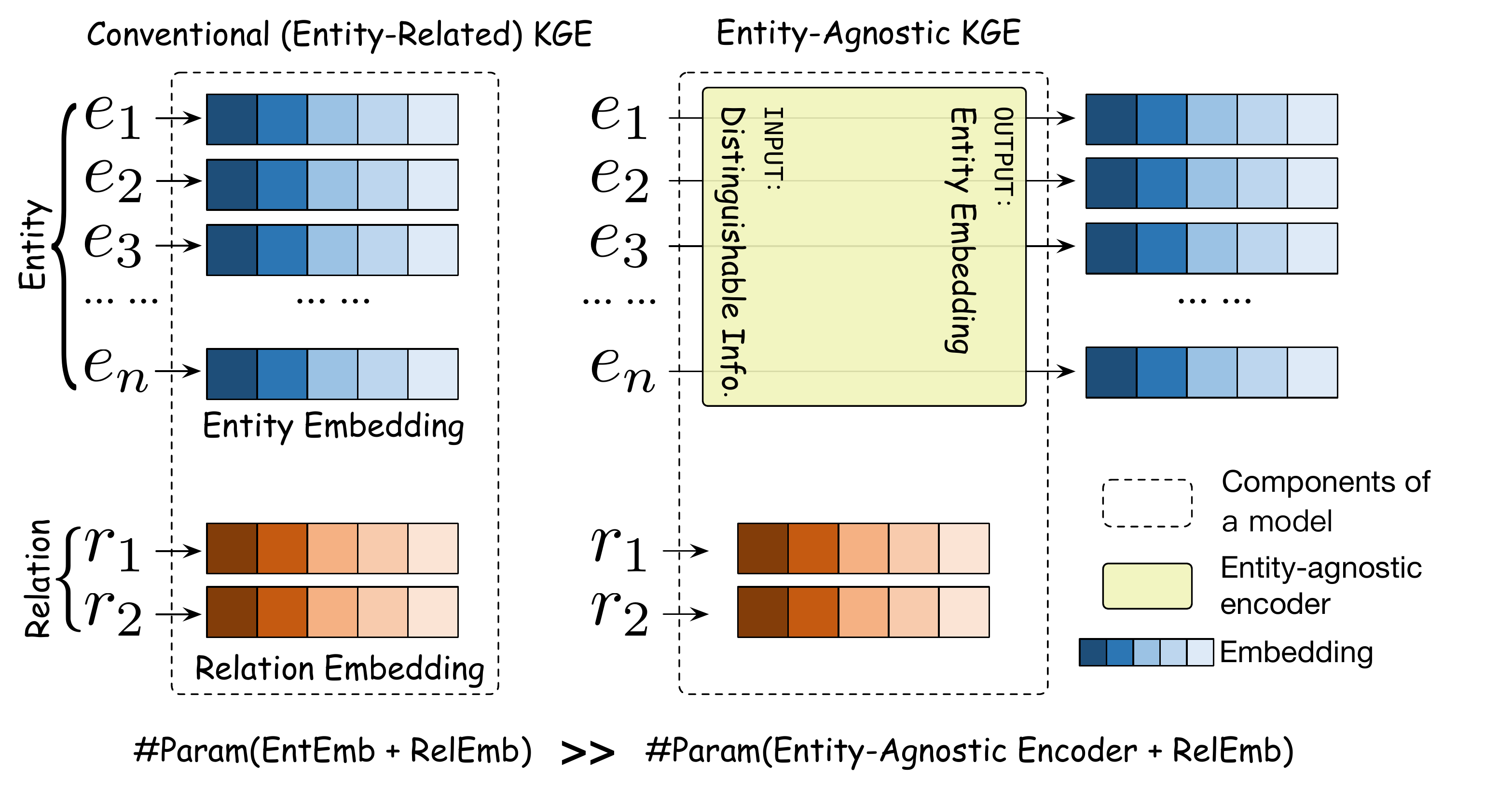}
\caption{A conventional KGE model (left) learns a specific embedding for each entity, and the model's components (i.e., entity embeddings) are related to entities. An entity-agnostic KGE model (right) learns an encoding approach for transforming entities' distinguishable information to their embeddings, avoiding maintaining a large embedding matrix.}
\label{fig:intro}
\end{figure}

Typically, knowledge graphs contain a large number of factual triples in the form of \textit{(head entity, relation, tail entity)}, or \textit{(h, r, t)} for short. A triple reflects a specific connection (i.e., relation) between two entities / concepts. 
However, since the incompleteness of KGs \cite{KnowPrompt}, a fundamental problem is knowledge graph completion \cite{TransE, RotatE,WPKD2020}.
Many knowledge graph embedding (KGE) methods \cite{KGSurvey} have been proposed and 
are theoretically and empirically shown to be effective forsolving the problem of knowledge graph completion. Conventionally, KGE methods map entities and relations from a KG into continuous vector spaces and predict missing links based on the computations of vector representations to complete knowledge graphs.

In KGE methods, entities and relations are often mapped into vectors with a specific dimension. 
For example, TransE \cite{TransE} maps both entities and relations to the same $d$-dimensional vector space, namely $\mathbb{R}^{d}$, and the total embedding matrix is in $\mathbb{R}^{(|\mathcal{E}| + |\mathcal{R}|) \times d}$, where $|\mathcal{E}|$ and $|\mathcal{R}|$ are number of entities and relations, respectively.
In practice, since $|\mathcal{E}|$ is much larger than $|\mathcal{R}|$, the number of embedding parameters scales linearly to the number of entities. As shown in Figure \ref{fig:intro}, we call conventional KGE methods \textit{entity-related KGE}, meaning that the components of a KGE model (i.e., entity embeddings) are related to entities.

Hence, the storage space costs for maintaining embeddings of entity-related KGE methods can be huge; e.g., 500-dimensional RotatE \cite{RotatE} maintains 123 million parameters for YAGO3-10 \cite{yago3-10}. 
Such inefficient linearly scaling space costs for entity-related KGE methods bring several challenges to real-world KG applications. 
For example, many deep learning models, including KGE models, are expected to be applied on edge devices \cite{MobileNets}, and colossal parameter space costs may weaken the feasibility.
Furthermore, some studies explore adapting federated learning \cite{FedAvg} to KGs and training KGE models decentralized \cite{FKGE, FedE}, and the number of parameters significantly increases communication costs in the federated learning scenario.
To this end, we argue that \textit{an entity-agnostic KGE method with a stable and relatively low parameter count, as well as independent of the number of entities}, is essential for solving the above issues and achieving efficient knowledge graph embedding.

In this paper, we propose a novel knowledge graph embedding method named \underline{E}ntity-\underline{A}gnostic \underline{R}epresentation \underline{L}earning, \textbf{\model}, in the sense that the components of \model~are \textit{not} mapped to entities, enabling the number of model parameters to not linearly scale up when the number of entities increases, as shown in Figure \ref{fig:intro}. 
Specifically, instead of learning a specific embedding for each entity as traditional KGE methods, we encode \textit{distinguishable information} of entities to represent them, and the encoding process is universal and \textit{entity-agnostic}.
First, in \model, we only learn embeddings for a small set of entities and refer to them as \textit{reserved entities}.
Then, we design the following three kinds of distinguishable information for encoding the embeddings for the full set of entities.
1) \textit{ConRel}: \underline{con}nected \underline{rel}ation information can make an entity distinguishable since the semantics of a relation's head or tail entities is often distinct and stable;
2) \textit{$k$NResEnt}: we also use the status of connected relation to retrieve \underline{$k$}-\underline{n}earest \underline{res}erved \underline{ent}ity information for an entity to improve the distinguishability;
3) \textit{MulHop}: we incorporate \underline{mul}ti-\underline{hop} neighbor information that different entities often have various ones. 
For the model design, we propose \textit{relational features} for entities to reflect the status of their connected relations for encoding ConRel and retrieving $k$NResEnt. Based on the above ConRel and $k$NResEnt encoding, a GNN is used to consider MulHop information and finally output embeddings for entities.

·We conduct an extensive empirical evaluation to show the effectiveness of our proposed \model. We train \model~on various KG benchmarks with different characteristics, and the results illustrate that we use fewer parameters and obtain better performance than baselines. The contributions of our work are summarised as follows:
\begin{itemize}
\item We point out the problem of entity-related KGE methods and emphasize the importance of exploring entity-agnostic representation learning for KGs.
\item We propose a novel KGE method, \model, which uses an entity-agnostic encoding process to encode entity embeddings based on their distinguishable information.
\item We conduct comprehensive experiments and show that our model is more parameter-efficient than baselines and achieves competitive performance.
\end{itemize}

\section{Related Work}
\subsection{Knowledge Graph Embedding}

For applying KGs to downstream tasks, including question answering \cite{QA-GNN, CQA,HGCL+2022}, search~\cite{PTT09}, recommendation \cite{recsys}, and some other in-KG tasks like link prediction \cite{TransE, IterE}, lots of studies are devoted to designing methods for mapping entities and relations of a KG into continuous vector spaces and remaining the inherent semantics in the KG. 
From the view of designing KGE models, they can be divided into various types \cite{kgembedding, neuralkg, KGSurvey}.

Conventional KGE methods are mainly categorized into translational distance models and semantic matching models. 
TransE \cite{TransE} is a classic translational distance model, which assumes that the relation is a translation vector from the head entity to the tail entity for a true triple. 
RotatE \cite{RotatE} defines the relation as a rotation between entities in the complex vector space and can capture various relation patterns.
For semantic matching models, DistMult \cite{DistMult} calculates the score for a triple by capturing the interactions between entity embeddings. ComplEx \cite{ComplEx} extend DistMult to map embeddings to the complex vector space for handling asymmetric relations.

Adapting graph neural networks (GNNs) to embed knowledge graphs has recently gained massive attention \cite{GEN, MorsE, RMPI}. Typically, R-GCN \cite{RGCN} is developed for multi-relational data and uses different transformation weights for various relations. CompGCN \cite{CompGCN} leverages entity-relation composition operations from score functions of other conventional KGE methods (e.g. TransE) for message passing. 

However, these   methods do not consider the efficiency of parameters, or data  compression~\cite{PPRW*2014,ZWHP2018} and simplification~\cite{WWTPA2014}   for KGs.  

\subsection{Parameter-Efficient Models}
 
With the increase of existing deep learning model sizes, reducing model parameters and making them more efficient has attracted much research, including network pruning \cite{DBLP:conf/iclr/MolchanovTKAK17}, quantification \cite{DBLP:conf/icml/LinTA16, DBLP:conf/acl/Sachan20}, parameter sharing \cite{DBLP:conf/iclr/DehghaniGVUK19, DBLP:conf/iclr/LanCGGSS20}, and knowledge distillation \cite{DBLP:journals/corr/HintonVD15}.


For knowledge graph embedding, methods based on quantization and knowledge distillation are studied more.
TS-CL \cite{DBLP:conf/acl/Sachan20} proposes a method based on quantization technology to reduce the size of KGEs by representing entities as vectors of discrete codes. 
LightKG \cite{DBLP:conf/cikm/WangWLG21} is a lightweight end-to-end KGE framework based on quantization, which contains a residual module to induce diversity among codebooks and proposes a novel dynamic negative sampling method based on quantization to further improve the performance of KGE.
MulDE \cite{DBLP:conf/www/Wang0MS21} applies the knowledge distillation technology to transfer the knowledge from multiple low-dimensional hyperbolic KGE teacher models to a student model. 
DualDE \cite{DBLP:conf/wsdm/ZhuZCCC0C22} considers the dual influence between the teacher and the student in the distillation process to make the teacher more suitable for the student and obtain better distillation results.

We do not directly compare against the methods above since they need to train standard KGE models in advance and then apply various compression methods. The most relevant work for our paper is NodePiece \cite{NodePiece}, a recently proposed compositional method for representing entities in KGs. It uses anchors and relations to encode entities with a fixed-size vocabulary.

\section{Methodology}

In the context of our work, a knowledge graph consists of an entity set $\mathcal{E}$, a relation set $\mathcal{R}$, and a triple set $\mathcal{T}$. More precisely, a knowledge graph is represented as $\mathcal{G} = (\mathcal{E}, \mathcal{R}, \mathcal{T})$, where $\mathcal{T} = \{(h,r,t)\} \subseteq \mathcal{E} \times \mathcal{R} \times \mathcal{E}$.
Conventional knowledge graph embedding methods often learn embeddings to represent every entity and relation to predict missing triples (namely link prediction) based on specific score functions \cite{TransE, RotatE}. 

\begin{figure}[t]
\centering
\includegraphics[scale=0.42]{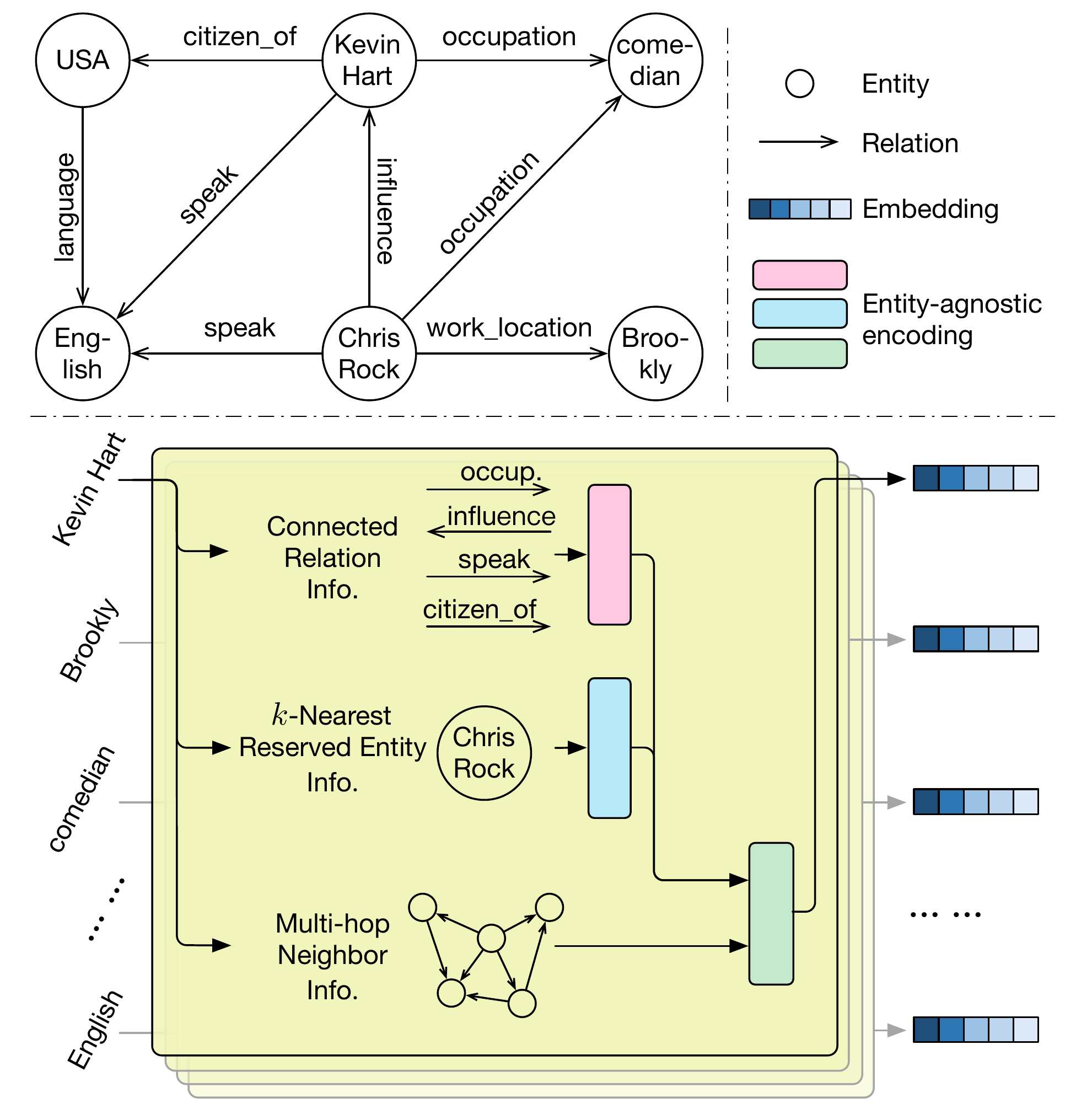}
\caption{Illustration of distinguishable information.}
\label{fig:dist-info}
\end{figure}

Rather than storing embeddings for all entities and relations, we aim to design a model with fewer parameters to encode entity embeddings and obtain competitive performance compared with conventional KGE methods to make parameters more efficient.
In our proposed entity-agnostic representation learning, EARL, we only learn specific embeddings for a small set of entities and refer to them as reserved entities $\mathcal{E}^{res}$. In practice, entities in $\mathcal{E}^{res}$ are randomly selected in advance.
We encode three kinds of distinguishable information to obtain embeddings for all entities. The encoding procedure can decrease the parameter space costs since the number of parameters in encoders is independent of the number of entities.
We show an intuition of this method in Figure \ref{fig:dist-info} and explain the details as follows.

Next, we first introduce the components and the overview of three distinguishable information in Section \ref{sec:dist-info}, then we describe the details of encoding such information to obtain entity embeddings in Section \ref{sec:ea-encoding}, and finally, we illustrate the training process of our model in Section \ref{sec:model-training}.

\begin{figure*}[t]
\centering
\includegraphics[scale=0.38]{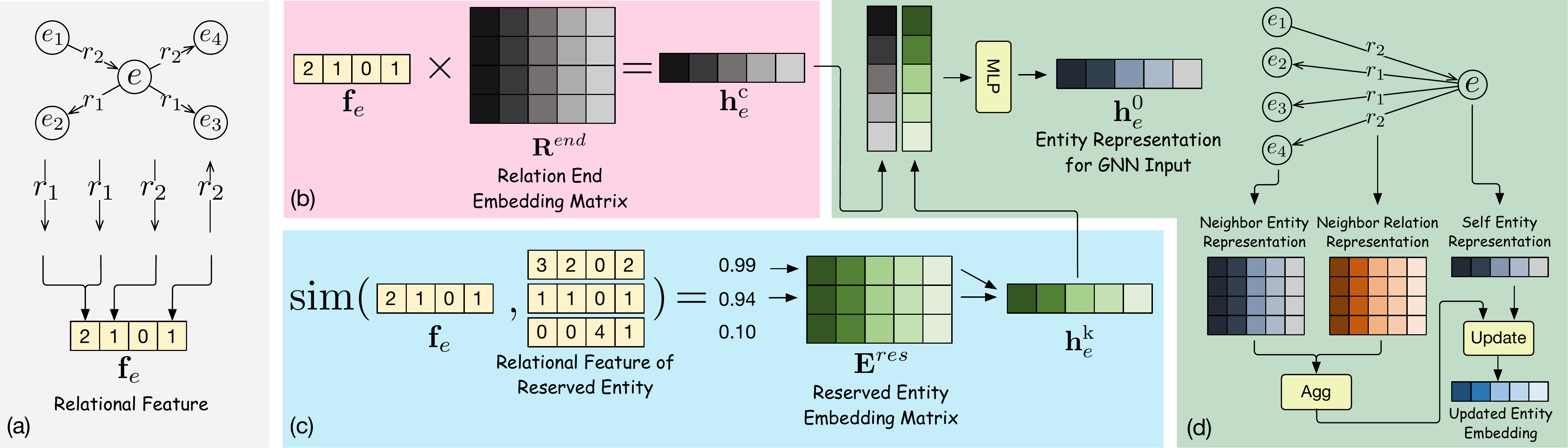}
\caption{Overview of (a) constructing relational features, (b) ConRel encoding, (c) $k$NResEnt encoding and (d) MulHop encoding.}
\label{fig:method}
\end{figure*}

\subsection{Distinguishable Information}
\label{sec:dist-info}

The goal of designing distinguishable information is to represent each entity as uniquely as possible. As shown in Figure \ref{fig:dist-info}, we give an example for a clear description of different distinguishable information. 

\subsubsection{ConRel.}
For a specific entity, we first use its connected relations, including their directions, as the connected relation information (ConRel) since different entities connect different relations, and connected relations can reflect entities' semantics~\cite{WCLXP+2021}. 
For example, head entities of the relation \texttt{occupation} are mainly people, and tail entities are mainly positions. 

\subsubsection{$k$NResEnt.}
Nevertheless, the information provided by relations may be unclear since the granularities of relation definitions may be varied across KGs. More precisely, a relation \texttt{film\_country} in Freebase \cite{freebase} reflects that it connects a film and a country. However, from a relation \texttt{hypernym} in WordNet \cite{WordNet}, we only know that it connects two words, while all the entities in WordNet are words.
Thus, connected relations (ConRel) cannot provide enough information to distinguish entities. 
Besides using relevant relations, we also use some relevant entities to represent a specific entity. We use $k$-nearest reserved entities ($k$NResEnt) as a kind of distinguishable information, and the similarity between entities is based on their connected relations. 
For example, in Figure \ref{fig:dist-info}, if the entity \texttt{Chris\_Rock} is a reserved entity and it is the nearest entity for \texttt{Kevin\_Hart}, then the embedding of \texttt{Chris\_Rock} will be used to encode the embedding of \texttt{Kevin\_Hart}.

\subsubsection{MulHop.}
Even though ConRel and $k$NResEnt can capture distinguishable information, they fail when two entities have the same relation connection. To enhance the distinguishing capability, we consider using multi-hop neighbor (MulHop) information since it is almost impossible for two entities to have both the same relation connection and multi-hop neighbor. Specifically, we use a GNN to update the above two distinguishable information for incorporating neighbor information from KG structures.

\subsection{Entity-Agnostic Encoding}
\label{sec:ea-encoding}

In \model, we only learn specific embeddings for a small set of entities (i.e., reserved entities) $\mathcal{E}^{res}$, and the embedding matrix of them is denoted as $\mathbf{E}^{res} \in \mathbb{R}^{|\mathcal{E}^{res}| \times d}$. These entities are randomly selected in advance, and their embeddings are trainable parameters. For the full set of entities, we obtain their embeddings via the following entity-agnostic encoding based on distinguishable information.

\subsubsection{ConRel Encoding.}
To formally represent the connected relations of an entity, we propose the \textit{relational feature}. Specifically, for each entity $e$, each dimension of its relational feature $\mathbf{f}_{e} \in \mathbb{Z}^{2|\mathcal{R}|}$ represents the frequency of being the head or the tail entity of a relation in $\mathcal{R}=\{r_{i}\}_{i=1}^{|\mathcal{R}|}$. Formally, we define each dimension of $\mathbf{f}_{e}$ as follows:
\begin{equation}
    \mathbf{f}_{e,i} =
    \left\{
    \begin{aligned}
    & {\rm H}(e, r_i),&  
    & i \leq |\mathcal{R}|\\
    & {\rm T}(e, r_{i-|\mathcal{R}|}),&
    & |\mathcal{R}| < i \leq 2|\mathcal{R}|,
    \end{aligned}
    \right.
\end{equation}
where ${\rm H}(e, r) = |\{(e,r,x)|\exists x,(e,r,x) \in \mathcal{T}\}|$ denotes the frequency of the entity $e$ as the head entity of the relation $r$; ${\rm T}(e, r) = |\{(x,r,e)|\exists x,(x,r,e) \in \mathcal{T}\}|$ denotes the frequency of the entity $e$ as the tail entity of the relation $r$. The visual illustration is shown in Figure \ref{fig:method}(a).

To maintain the semantics of being head or tail entities of relations, we propose \textit{relation end embeddings} $\mathbf{R}^{end} \in \mathbb{R}^{2|\mathcal{R}| \times d}$ to encode relational feature:
\begin{equation}
\mathbf{h}_{e}^{\rm c} = f_{r}(\mathbf{f}_{e}^{\top} \mathbf{R}^{end}),
\end{equation}
where $f_{r}: \mathbb{R}^{d} \rightarrow \mathbb{R}^{d}$ is a 2-layer MLP; $\mathbf{h}_{e}^{\rm c}$ denotes the encoded ConRel information for entity $e$. 

\subsubsection{$k$NResEnt Encoding.}

For encoding the information from $k$-nearest reserved entities for the entity $e$, we first calculate the cosine similarity between $e$ and each entity $e_i$ in $\mathcal{E}^{res}$ based on relational features:
\begin{equation}
{\rm sim}(e, e_{i}) = \frac{\mathbf{f}_{e}^{\top}\mathbf{f}_{e_{i}}}{\|\mathbf{f}_{e}\|\|\mathbf{f}_{e_{i}}\|}.
\end{equation}
Next, we retrieve the top-$k$ reserved entities based on similarity values:
\begin{equation}
\mathcal{P}^{k}_{e} = {\rm Top}^{k}(\{{\rm sim}(e, e_{i})|e_{i} \in \mathcal{E}^{res} \}),
\end{equation}
where $\mathcal{P}^{k}_{e}$ is a top-$k$ reserved entity set for the entity $e$, and $k$ is a hyper-parameter.

For utilizing the retrieved reserved entities, we use a weighted sum to encode $k$-nearest reserved entity information as follows:
\begin{equation}
\begin{aligned}
\mathcal{V}^{k}_{e} &= {\rm Softmax}(\{{\rm sim}(e, e_{i})|e_{i} \in \mathcal{P}^{k}_{e} \}),\\
\mathbf{h}_{e}^{\rm k} &= \sum_{e_i \in \mathcal{P}^{k}_{e}, v_i \in \mathcal{V}^{k}_{e},}  v_i \mathbf{E}^{res}_{e_i},
\end{aligned}
\end{equation}
where $\mathbf{E}^{res}_{e_i}$ denotes the embedding for the reserved entity $e_i$, and $\mathbf{h}_{e}^{\rm k}$ denotes the encoded $k$NResEnt information for entity $e$.

\subsubsection{MulHop Encoding.}

For incorporating multi-hop neighbor information, we use a GNN to update $\mathbf{h}_{e}^{\rm c}$ and $\mathbf{h}_{e}^{\rm k}$ for each entity $e$. We use the above two kinds of encoded information as the input representations of GNN, and we combine them as follows: 
\begin{equation}
    \mathbf{h}_{e}^{0} = f_{m}([\mathbf{h}_{e}^{\rm c}; \mathbf{h}_{e}^{\rm k}]),
\label{eq:info-combine}
\end{equation}
where the operation $[\cdot; \cdot]$ denotes the vector concatenation, and $f_{m}: \mathbb{R}^{2d} \rightarrow \mathbb{R}^{d}$ is a 2-layer MLP. 
Furthermore, $\mathbf{h}_{e}^{0}$ is the input representation of the GNN for $e$. Note that for the non-reserved entities, we use Equation (\ref{eq:info-combine}) to obtain the input representation; for the reserved entities in $\mathcal{E}^{res}$, their input representations are directly looked up from the embedding matrix $\mathbf{E}^{res}$.

In our GNN framework, similar to previous works \cite{CompGCN, MaKEr}, we use a linear transformation on the concatenation of entity and relation representations to aggregate the neighbor information. Specifically, the message aggregation for the entity $e$ is:
\begin{equation}
\begin{aligned}
    \mathbf{m}_{e}^{l} = \sum_{(r, t) \in \mathcal{O}(e)} \mathbf{W}_{\text{out}}^l [\mathbf{h}^l_r; \mathbf{h}^l_t] + \sum_{(r,h) \in \mathcal{I}(e)} \mathbf{W}_{\text{in}}^l [\mathbf{h}^l_r; \mathbf{h}^l_h],
\label{eq:gnn-agg}
\end{aligned}
\end{equation}
where $\mathcal{O}(e)$ denotes the out-going relation-entity pair set of $e$ and $\mathcal{I}(e)$ denotes the in-going relation-entity pair set. $\mathbf{W}_{\text{out}}^l$ and $\mathbf{W}_{\text{in}}^l$ are transformation matrices for out-going and in-going pairs. $l \in [0, \dots, L]$ denotes the layer of GNN and $L$ is the total number of GNN layers. The input entity representations are calculated in Equation (\ref{eq:info-combine}), and the input relation representations (e.g., $\mathbf{h}_{r}^{0}$) are looked up in a trainable relation embedding matrix $\mathbf{R} \in \mathbb{R}^{|\mathcal{R}|\times d}$. 

The entity representation of $e$ in the GNN is updated as follows:
\begin{equation}
    \mathbf{h}_{e}^{l+1} = \sigma \left( \frac{1}{c}\mathbf{m}_{e}^{l} + \mathbf{W}_{\text{self}}^{l} \mathbf{h}_{e}^{l} \right),
\label{eq:gnn-update}
\end{equation}
where $c=|\mathcal{I}(e)+\mathcal{O}(e)|$ is a normalization constant. $\mathbf{W}_{\rm self}^{l}$ is a matrix for self representation update, and $\sigma$ is an activation function. Furthermore, relation representations will also be updated in each layer: $\mathbf{h}_{r}^{l+1} = \sigma \left( \mathbf{W}_{\text{rel}}^{l} \mathbf{h}_{r}^{l} \right)$. 
We use the output representations in the $L$-th layer for entities and relations as their embeddings to calculate scores next.

\subsection{Model Training}
\label{sec:model-training}

Following the conventional KGE training regime, we optimize \model~to score true triples in the training set higher than sampled negative triples. 
Many score functions can be used in \model.
To show its versatility, we use RotatE \cite{RotatE}, a representative method and one of the state-of-the-art KGE models, as the score function in \model: $f(h,r,t) = -\| \mathbf{h} \circ \mathbf{r} - \mathbf{t} \|$.
Here entities and relations are mapped into complex vector spaces, $\mathbf{h}, \mathbf{r}, \mathbf{t} \in \mathbb{C}^d$. 

As for the loss function, we apply a widely used self-adversarial negative sampling loss:
\begin{equation}
\begin{aligned}
    \mathcal{L}(h,r,t) = &
    - \log \sigma\left(
        \gamma + f(h,r,t)
    \right) \\
    &-\sum_{i=1}^{n} p\left(h_{i}^{\prime}, r, t_{i}^{\prime}\right) \log \sigma\left(
        -\gamma-f(h_{i}^{\prime},r,t_{i}^{\prime})
    \right),
\end{aligned}
\end{equation}
where $\gamma$ is a fixed margin and $\sigma$ is the sigmoid function. $(h_{i}^{\prime},r,t_{i}^{\prime})$ is a sampled negative triple for $(h,r,t)$ and $n$ is the number of negative triples. $p\left(h_{i}^{\prime}, r, t_{i}^{\prime}\right)$ is the self-adversarial weight for this negative triple, and the calculation of this weight is as follows:
\begin{equation}
    p\left(h_j^{\prime}, r, t_{j}^{\prime}\right) = \frac{\exp \alpha f(h_j^{\prime}, r, t_{j}^{\prime})}{\sum_{i} \exp \alpha f(h_i^{\prime}, r, t_{i}^{\prime})}, 
\label{eq:adv-sample}
\end{equation}
where $\alpha$ is the temperature factor.

\section{Experiments}

In this section, we conduct extensive experiments and analyses on various datasets to show the effectiveness of our proposed \model. Note that the focus of our model is not outperforming the state-of-the-art KGE methods but showing that we are more parameter-efficient. Thus, this section is motivated by the following research questions: (\textbf{RQ1}) Is \model~parameter-efficient and capable of obtaining competitive performance? (\textbf{RQ2}) How does the effectiveness of components in \model~on different datasets? (\textbf{RQ3}) What is the impact of different settings on \model? The source code is available at \url{https://github.com/zjukg/EARL}.

\subsection{Experimental Setting}

\subsubsection{Datasets and Baselines.}

\begin{table}[t]
\centering
\resizebox{\columnwidth}{!}{
\begin{tabular}{lrrrrr}
\toprule
Dataset & \#Ent &\#Rel & \#Train & \#Valid & \#Test \\
\midrule
FB15k-237 & 14,505 & 237 & 272,115 & 17,526 & 20,438 \\
WN18RR & 40,559 & 11 & 86,835 & 2,824 & 2,924 \\
CoDEx-L & 77,951 & 69 & 551,193 & 30,622 & 30,622 \\
YAGO3-10 & 123,143 & 37 & 1,079,040 & 4,978 & 4,982 \\
\bottomrule
\end{tabular}
}
\caption{Dataset statistics. The number of entities, relations, training triples, validation triples, and test triples.}
\label{tab:statistic}
\end{table}

Our model is evaluated on several KG benchmarks with various sizes and characteristics, and the dataset statistics are shown in Table \ref{tab:statistic}. Specifically, FB15k-237 \cite{fb15k237} is derived from Freebase \cite{freebase} with 237 relations, and the inverse relations are deleted. WN18RR \cite{ConvE} is a subset of WordNet \cite{WordNet} with inverse relations deleted. CoDEx \cite{CoDEx} is a recently proposed KG benchmark that contains more diverse and interpretable content and is more difficult than previous datasets. CoDEx-L is the large-size version. YAGO3-10 \cite{yago3-10} is a subset of YAGO3, which consists of entities that have a minimum of 10 relations each.

For comparison, we use RotatE \cite{RotatE}, a representative and one of the state-of-the-art KGE methods, as a baseline, and the number of its model parameters can be controlled by the embedding dimension. 
Moreover, NodePiece \cite{NodePiece} which uses anchor nodes and node tokenization is the most proper baseline for \model.

\subsubsection{Evaluation Metrics.}

We evaluate models by the performance of link prediction on KGs, namely predicting missing triples in test sets. We report Mean Reciprocal Rank (MRR) and Hits@10 in the filtered setting \cite{TransE}. 
For quantifying the efficiency of models, we propose a metric calculated by \textit{MRR/\#P} (\#P denotes the number of parameters), and we denote it as \textit{Effi}.
For a fair comparison with baselines, we don't test the triples which involve entities that do not appear in the corresponding training sets.

\subsubsection{Implementation Details.}

We conduct our experiments on NVIDIA RTX 3090 GPUs with 24GB RAM, and we use PyTorch \cite{pytorch} and DGL \cite{DGL} for handling automatic differentiation and graph structure modeling. 
For entity-agnostic encoding, we use 2-layer GNNs, and the default number of $k$ for $k$NResEnt encoding is 10. 
We set the number of reserved entities as 10\% of the number of all entities for each dataset, namely 1450, 4055, 7795, and 12314 for FB15k-237, WN18RR, CoDEx-L, and YAGO3-10.
For model training, the learning rate is set to 0.001; the batch size is set to 1024; the number of negative samples (i.e., $n$) is set to 256; the margin is set to 15 for YAGO3-10 and 10 for other datasets.

\subsection{Main Results}

\begin{table*}[t]
\centering
\resizebox{\textwidth}{!}{
\begin{tabular}{lccccc|ccccc}
\toprule
& \multicolumn{5}{c}{\textbf{FB15k-237}} & \multicolumn{5}{c}{\textbf{WN18RR}} \\
\cmidrule(lr){2-6} \cmidrule(lr){7-11}
& \multicolumn{1}{c}{{Dim}} &\multicolumn{1}{c}{{\#P(M)}} & 
\multicolumn{1}{c}{{MRR}} & \multicolumn{1}{c}{{Hits@10}} & 
\multicolumn{1}{c}{{Effi}} & 
\multicolumn{1}{c}{{Dim}} & \multicolumn{1}{c}{{\#P(M)}} & 
\multicolumn{1}{c}{{MRR}} & \multicolumn{1}{c}{{Hits@10}} &
\multicolumn{1}{c}{{Effi}} \\ 
\midrule

RotatE
& 1000 & 29.3 & 0.336 & 0.532 & 0.011 
& 500 & 40.6 & 0.508 & 0.612 & 0.013 \\ 

\rowcolor{blue!10}
RotatE
& 100 & 2.9 & 0.296 & 0.473 & 0.102  
& 50 & 4.1 & 0.411 & 0.429 & 0.100 \\

\midrule

\rowcolor{blue!10}
NodePiece + RotatE *
& 100 & 3.2 & 0.256 & 0.420 & 0.080 
& 100 & 4.4 & 0.403 & 0.515 & 0.092 \\

\midrule

\rowcolor{red!10}
\model~+ RotatE
& 150 & 1.8 & 0.310 & 0.501 & 0.172   
& 200 & 3.8 & 0.440 & 0.527 & 0.116  \\

w/o Reserved Entity
& 150 & 1.1 & 0.306 & 0.492 & 0.278 
& 200 & 1.7 & 0.347 & 0.461 & 0.204  \\

w/o ConRel
& 150 & 1.2 & 0.309 & 0.501 & 0.257 
& 200 & 3.0 & 0.432 & 0.520 & 0.144  \\

w/o $k$NResEnt
& 150 & 1.6 & 0.301 & 0.488 & 0.188 
& 200 & 3.3 & 0.409 & 0.498 & 0.124  \\

w/o ConRel + $k$NResEnt
& 150 & 1.2 & 0.302 & 0.486 & 0.251  
& 200 & 3.0 & 0.350 & 0.479 & 0.117  \\

w/o MulHop
& 150 & 1.1 & 0.250 & 0.414 & 0.227   
& 200 & 2.4 & 0.048 & 0.084 & 0.020  \\

\bottomrule
\end{tabular}
}
\caption{Link prediction results on FB15k-237 and WN18RR. Results of * are taken from \citet{NodePiece}}
\label{tab:lp-fb-wn}
\end{table*}

\begin{table*}[t]
\centering
\resizebox{\textwidth}{!}{
\begin{tabular}{lccccc|ccccc}
\toprule
& \multicolumn{5}{c}{\textbf{CoDEx-L}} & \multicolumn{5}{c}{\textbf{YAGO3-10}} \\
\cmidrule(lr){2-6} \cmidrule(lr){7-11}
& \multicolumn{1}{c}{{Dim}} &\multicolumn{1}{c}{{\#P(M)}} & 
\multicolumn{1}{c}{{MRR}} & \multicolumn{1}{c}{{Hits@10}} & 
\multicolumn{1}{c}{{Effi}} & 
\multicolumn{1}{c}{{Dim}} & \multicolumn{1}{c}{{\#P(M)}} & 
\multicolumn{1}{c}{{MRR}} & \multicolumn{1}{c}{{Hits@10}} &
\multicolumn{1}{c}{{Effi}} \\ 
\midrule

RotatE *
& 500 & 78.0 & 0.258 & 0.387 & 0.003 
& 500 & 123.2 & 0.495 & 0.670 & 0.004 \\

\rowcolor{blue!10}
RotatE *
& 25 & 3.8 & 0.196 & 0.322 & 0.052 
& 20 & 4.8 & 0.121 & 0.262 & 0.025 \\

\midrule

\rowcolor{blue!10}
NodePiece + RotatE *
& 100 & 3.6 & 0.190 & 0.313 & 0.053 
& 100 & 4.1 & 0.247 & 0.488 & 0.060 \\

\midrule

\rowcolor{red!10}
\model~+ RotatE
& 100 & 2.1 & 0.238 & 0.390 & 0.113
& 100 & 3.0 & 0.302 & 0.498 & 0.101 \\

w/o Reserved Entity
& 100 & 0.5 & 0.203 & 0.337 & 0.406 
& 100 & 0.4 & 0.119 & 0.226 & 0.296 \\

w/o ConRel
& 100 & 1.9 & 0.237 & 0.384 & 0.124 
& 100 & 2.8 & 0.322 & 0.522 & 0.115 \\

w/o $k$NResEnt
& 100 & 2.0 & 0.232 & 0.374 & 0.116 
& 100 & 2.9 & 0.249 & 0.429 & 0.086 \\

w/o ConRel + $k$NResEnt
& 100 & 1.9 & 0.234 & 0.375 & 0.123
& 100 & 2.8 & 0.286 & 0.487 & 0.102 \\

w/o MulHop
& 100 & 1.8 & 0.095 & 0.174 & 0.053
& 100 & 2.7 & 0.033 & 0.048 & 0.012 \\

\bottomrule
\end{tabular}
}
\caption{Link prediction results on CoDEx-L and YAGO3-10. Results of * are taken from \citet{NodePiece}}
\label{tab:lp-codex-yago}
\end{table*}

We summarize the results on four datasets in Table \ref{tab:lp-fb-wn} and Table \ref{tab:lp-codex-yago}. 
We report the results of RotatE with large parameter counts to show the approximate upper-bound performance on datasets. Moreover, results from RotatE with similar parameter counts as NodePiece and EARL are used for comparison.
Note that we don't try to outperform conventional KGE methods on a large parameter budget since parameter efficiency is not an important factor in that scenario but performance.

From these results, comparing the performance of EARL (in \colorbox{red!10}{red}) with NodePiece and RotatE using a similar parameter budget (in \colorbox{blue!10}{blue}), we find that \model~outperforms them on MRR and Hits@10 while using fewer parameters.
Specifically, on FB15k-237, \model~uses only 62\% parameters and obtains a relative increase of 4.7\% on MRR in comparison with RotatE. On WN18RR, \model~achieves a 7\% MRR improvement with 93\% parameters compared with RotatE. In these two datasets, another baseline NodePiece uses even more parameters but does not 
outperform RotatE on MRR. 
On CoDEx-L, \model~uses 58\% parameters and increases 25.4\% relatively on MRR compared with NodePiece. \model~improves MRR with 22.2\% on YAGO3-10 with only 73\% parameters in contrast with NodePiece.

\begin{figure}[t]
\centering
\includegraphics[width=\columnwidth]{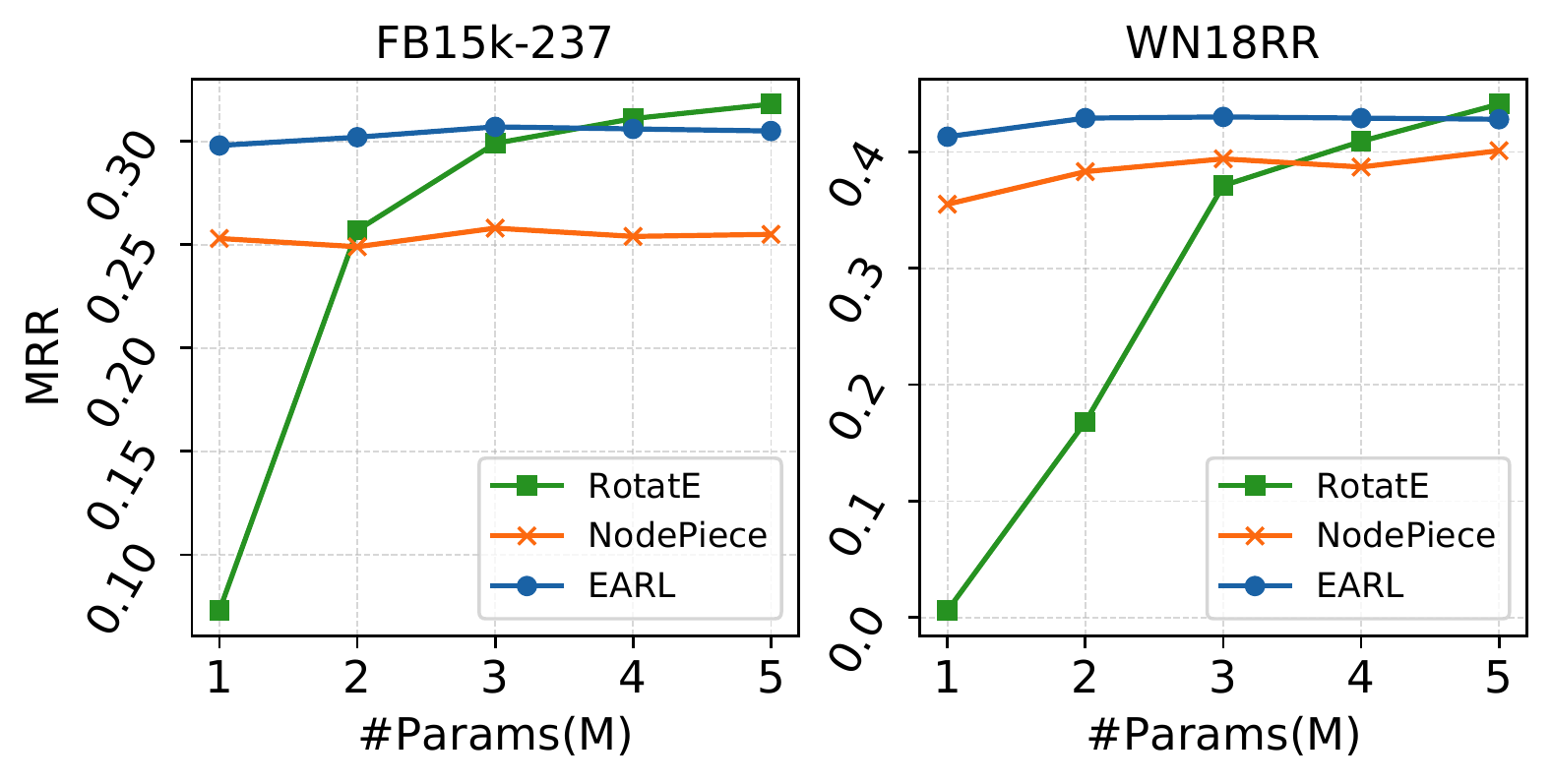}
\caption{Performance with different parameter budgets.}
\label{fig:same-param}
\end{figure}

Furthermore, apart from analyzing the effectiveness of \model~from static parameter budgets in tables, we also explore the performance of different models with dynamic parameter counts in Figure \ref{fig:same-param}. We adjust the dimensions of models to make their number of parameters from 1 to 5 million. We find that \model~obtains stable performance no matter how much the parameter is and improves significantly compared with RotatE in relatively low parameter budgets. 
Even though the results from the parameter-efficient baseline NodePiece are also stable, \model~achieves steady improvements in contrast with it. 
The dramatic drop with parameter decrease for RotatE's performance shows the inefficiency of entity-related KGE on small parameter budgets.

Finally, from the values of metric Effi in Table \ref{tab:lp-fb-wn} and \ref{tab:lp-codex-yago}, it's intuitive that \model~is more parameter-efficient than baselines. Above results indicate that our proposed \model~is parameter-efficient and answer the \textbf{RQ1}.

\subsection{Ablation Study}


In this part, we further probe the roles of different components in \model~by removing them separately. There are five types of ablation studies, as shown in Table \ref{tab:lp-fb-wn} and \ref{tab:lp-codex-yago}. More precisely, for ``w/o Reserved Entity", there are no reserved entities, and naturally, the $k$NREnt information is also disabled. For ``w/o ConRel + $k$NResEnt", we use random entity representations for the GNN to encode MulHop information and output entity embeddings. Moreover, ``w/o ConRel", ``w/o $k$NResEnt" and ``w/o MulHop" remove corresponding distinguishable information, respectively. Since the different characteristics of different datasets, the trends of ablation study results are distinct. We analyze them as follows respectively.

For \textit{FB15k-237}, except for removing MulHop, other ablation settings affect the performance slightly but not significantly.
For \textit{WN18RR}, ``w/o Reserved Entity" and ``w/o $k$NResEnt" impairs the performance. Replacing ConRel and $k$NResEnt with random representations (``w/o ConRel + $k$NResEnt") also affect the results. Moreover, the performance is affected dramatically by removing MulHop information. 
For \textit{CoDEx-L}, the trend is similar to that of FB15k-237, and ``w/o MulHop" has a remarkable influence on performance.
For \textit{YAGO3-10}, the trend is similar to that of WN18RR.

From the above analysis, we find that in ablation studies, FB15k-237 and CoDEx-L have a similar trend, and WN18RR and YAGO3-10 have a similar trend. We explain this from their data statistics. That is, FB15k-237 and CoDEx-L have more relations than WN18RR and YAGO3-10, 
and diverse relations provide enough distinguishable information for entity embeddings. Thus, even in the ``w/o Reserved Entity" and ``w/o $k$NResEnt", performance is not affected dramatically since ConRel information still exists.

Overall, the ablation study shows the effectiveness of components of \model~and the different ablation behaviors on datasets with different statistics, which answers the \textbf{RQ2}.

\subsection{Further Analysis}


\begin{figure}[t]
\centering
\includegraphics[width=\columnwidth]
{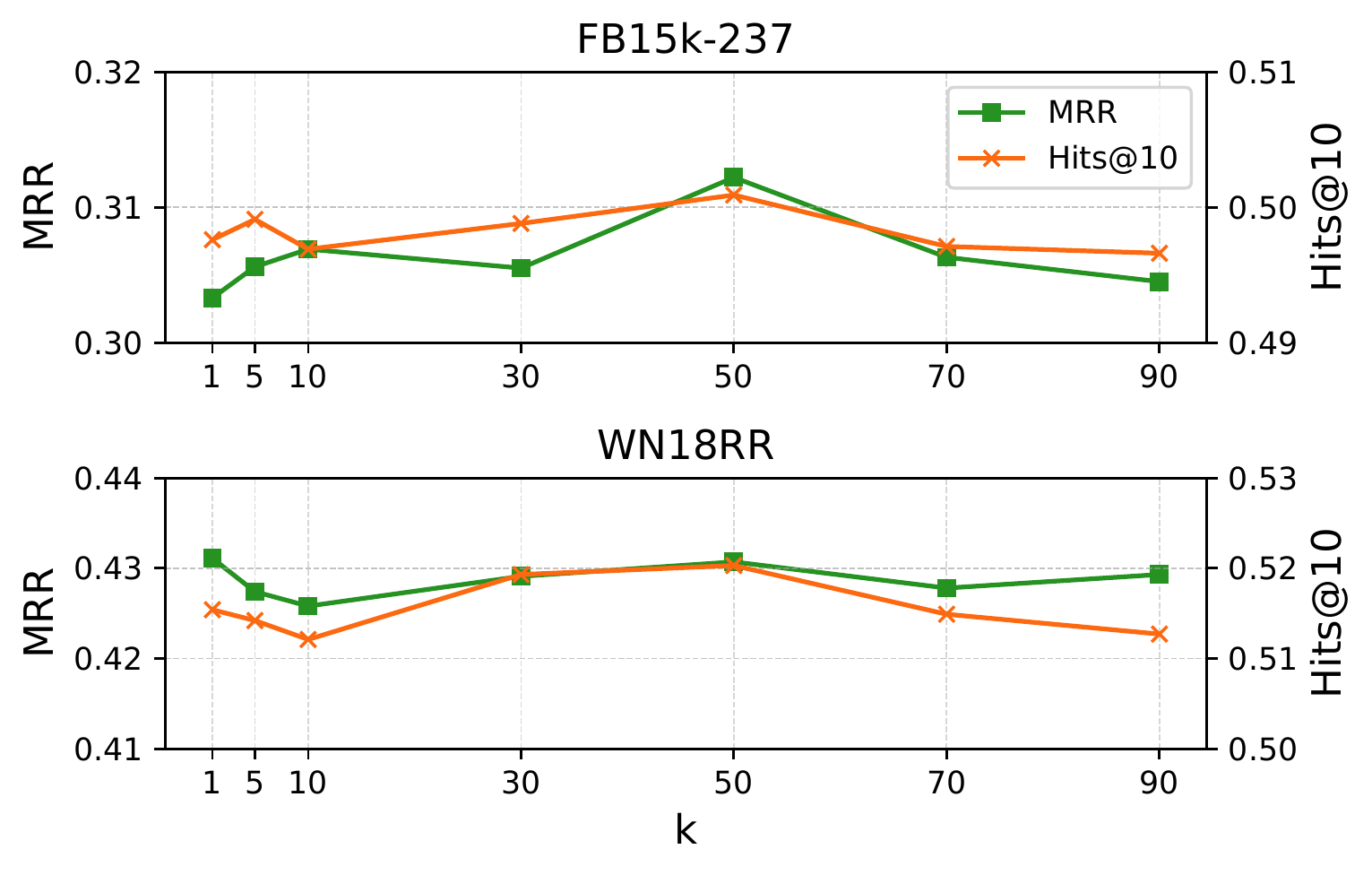}
\caption{Performance with various $k$.}
\label{fig:topk}
\end{figure}

\subsubsection{Analysis of $k$.} We also investigate the effect of the value of $k$ for retrieving $k$-nearest reserved entities to encode $k$NResEnt information. Based on FB15k-237 and WN18RR, we plot the MRR and Hits@10 results of training \model~with $k=[1,5,10,30,50,70,90]$ with 14,000 and 20,000 steps respectively in Figure \ref{fig:topk}. 
We find that our proposed \model~is robust to the value of $k$ since the variance in MRR and Hits@10 is about 0.01, and there is no significant performance change. 
Specifically, we observe that $k$ in a middle value (e.g., 50) leads to better performance, indicating that the results don't improve as $k$ increases. 
It is possible that a higher $k$ brings more noise into training. Moreover, a higher $k$ costs more training computation in practice. 

\subsubsection{Fixed Parameter Budget.} 

\begin{figure}[t]
\centering
\includegraphics[width=\columnwidth]
{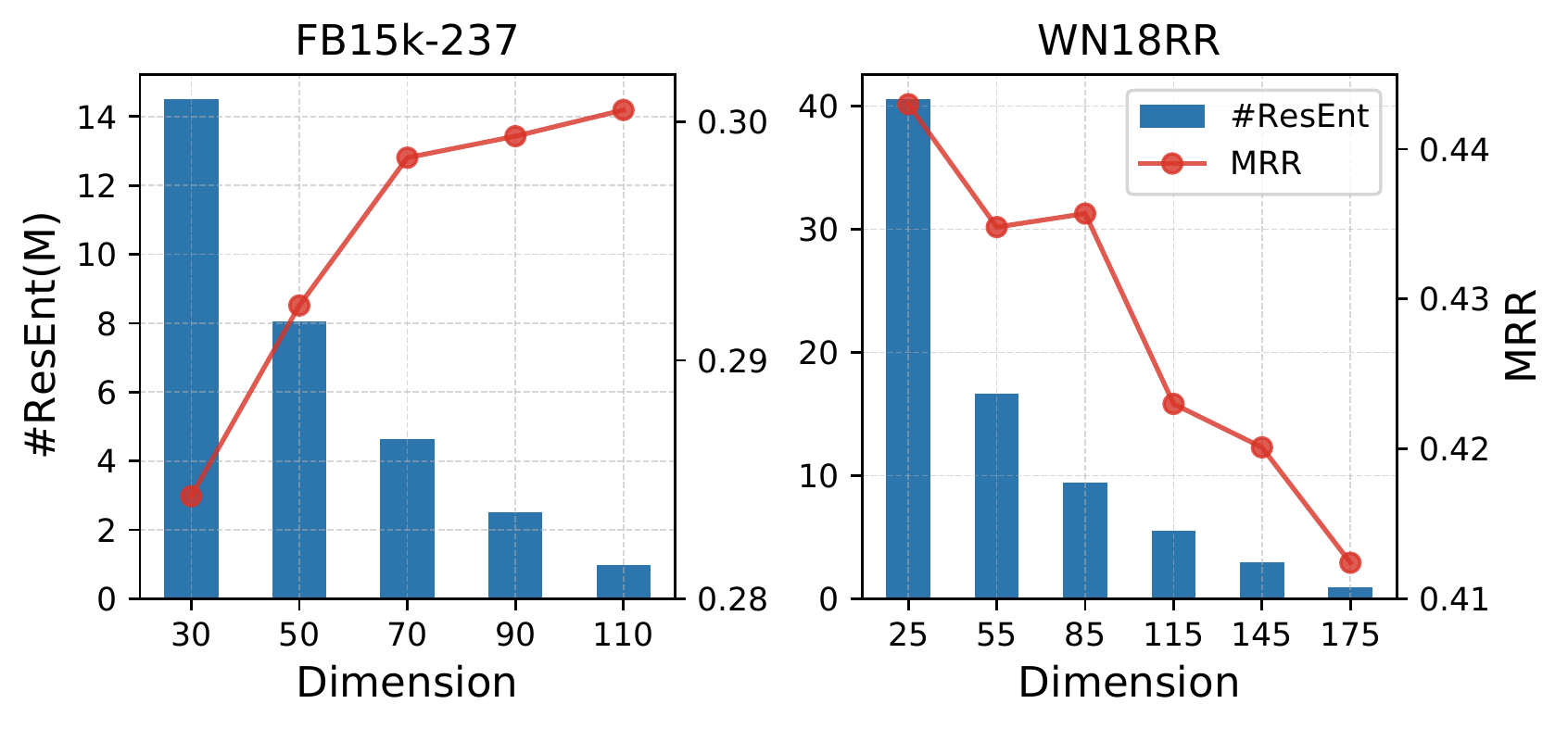}
\caption{Performance of different model settings for a fixed parameter budget.}
\label{fig:fix_param}
\end{figure}

Two main factors controlling the parameter count in \model~are the embedding dimension and the number of reserved entities (\#ResEnt), which contribute to the flexibility of \model~since we can adjust different values on the dimension and the \#ResEnt for a fixed parameter budget.
Given the 1M and 2M parameter budgets on FB15k-237 and WN18RR respectively, we slide the dimensions and adjust \#ResEnt to satisfy the parameter budgets. 
From Figure \ref{fig:fix_param}, we find that on FB15k-237, dimension is a more critical factor in influencing the performance since MRR improves as the increase of dimension. On WN18RR, \#ResEnt is more important. As the increase of \#ResEnt, even though the dimension is very small (i.e., 25), \model~obtains better performance on WN18RR. 
These results are consistent with our analyses in ablation studies that FB15k-237 has more diverse relations for encoding entities discriminatively, while WN18RR depends more on reserved entities for distinguishable information.
We suggest that when using \model~with a fixed parameter budget, the dimension and \#ResEnt can be adjusted for better performance based on datasets' characteristics.
Above analyses on various settings of \model~finally answer \textbf{RQ3}.

\section{Conclusion}

In this paper, we propose an entity-agnostic representation learning framework, \model, for achieving parameter-efficient knowledge graph embedding.
Unlike conventional entity-related KGE methods, \model~is entity-agnostic and does not map the model components to entities, preventing the number of parameters from scaling up linearly as the number of entities increases. 
Specifically, we design three kinds of distinguishable information to represent entities and then use an entity-agnostic encoding process to encode entity embeddings.
Extensive empirical results show the effectiveness of our embedding encoding process and the parameter efficiency of \model.


\section*{Acknowledgements}

This work is partially supported by NSFC U19B2027 and 91846204, with Mingyang Chen supported by the  China Scholarship Council (No. 202206320309) and Jeff Z. Pan supported by the Chang Jiang Scholars Program (J2019032).

\bibliography{aaai23}

\end{document}